\documentclass{article}
\usepackage{spconf,amsmath,graphicx}
\usepackage{graphicx}
\usepackage{threeparttable}
\usepackage{multirow}
\usepackage{times}
\usepackage{soul}
\usepackage{url}
\usepackage[utf8]{inputenc}  
\usepackage{graphicx}
\usepackage{amsmath}
\usepackage{amsthm}
\usepackage{booktabs}
\usepackage{algorithm}
\usepackage{algorithmic}
\usepackage[switch]{lineno}
\usepackage{subfigure}
\usepackage{rotating}
\newcommand{\tabincell}[2]{\begin{tabular}{@{}#1@{}}#2\end{tabular}}

\title{BOOSTING PRUNED NETWORKS WITH LINEAR OVER-PARAMETERIZATION}

\name{Yu Qian$^{1*}$, Jian Cao$^{1}$\sthanks{Equal contribution.}\sthanks{Corresponding author.} , Xiaoshuang Li$^{3*}$, Jie Zhang$^{2}$, Hufei Li$^{1}$, Jue Chen$^{1}$ }
  
\address{$^{1}$ Peking University,
  $^{2}$ Zhejiang University,
  $^{3}$ Shanghai Jiao Tong University
  }

\begin{document}
% \ninept
%
\maketitle
\begin{abstract}
Structured pruning is a popular technique for reducing the computational cost and memory footprint of neural networks by removing channels. 
It often leads to a decrease in network accuracy, which can be restored through fine-tuning.  
% However, as the pruning ratio grows larger, it becomes progressively more difficult to restore the accuracy of the trimmed network.
However, as the pruning ratio increases, it becomes progressively more difficult to restore the accuracy of the trimmed network.
% To overcome the challenge of restoring accuracy, we propose a novel method that linearly over-parameterizes the compact layers in the pruned network and then re-parameterizes them to their original layers after fine-tuning.
% To overcome these challenges, we propose a method that linearly over-parameterizes the compact layers in the pruned network and then re-parameterizes them to their original layers after fine-tuning. 
% To overcome these challenges, we propose a method that linearly over-parameterizes the compact layers in the pruned network which increases the number of learnable parameters contributing to better accuracy restoration. And then re-parameterizes them to their original layers after fine-tuning. 
To overcome this challenge, we propose a method that linearly over-parameterizes the compact layers, increasing the number of learnable parameters, and re-parameterizes them back to their original layers after fine-tuning, therefore contributing to better accuracy restoration. 
% We also use similarity-preserving knowledge distillation to encourage the over-parameterized block to learn the immediate data-to-data similarities of the corresponding dense layer. 
% Similarity-preserving knowledge distillation is exploited to encourage the over-parameterized block to learn the immediate data-to-data similarities of the corresponding dense layer. 
Similarity-preserving knowledge distillation is exploited to maintain the feature extraction ability of the expanded layers. 
% 同样的，Similarity-preserving knowledge distillation这个创新点/方法是否对应上面的challenge？它鼓励了the immediate data-to-data similarities，所以怎么清楚的描述这个特点帮到了restoring accuracy呢？是reviewer会觉得欠那么一点意思没讲清楚的。分配的reviewer很有可能不是特定领域能完全看得懂这些名词。
% Our method can be easily integrated with existing pruning techniques without additional technology, enhancing the pruning process for most existing networks.
Our method can be easily integrated without additional technology, making it compatible and can effectively benefit the pruning process for existing pruning techniques.
% Our method can be easily integrated with existing pruning techniques without additional technology, %描述方法可插入性很强，这里可以加一句这个便携插入特点能给我们的方法带来什么优势。比如make it portable/compatible and can effectively benefit the pruning process of most existing networks类似，可以再想一想
% Similarity-preserving knowledge distillation is leveraged to ensure output consistency which preserves the feature extraction capability of the expanded layers.
% Our experimental results on CIFAR-10 and ImageNet demonstrate that our approach outperforms the vanilla fine-tuning approach under different pruning strategies, particularly for large pruning ratios. 
% Our approach outperforms the vanilla fine-tuning on CIFAR-10 and ImageNet under different pruning strategies, particularly with large pruning ratios. 
Our approach outperforms fine-tuning on CIFAR-10 and ImageNet under different pruning strategies, particularly when dealing with large pruning ratios. 
\end{abstract}
\section{Introduction}
\label{sec:intro}

%Network pruning \cite{ding2021resrep,gao2021network,liang2021pruning,wang2021accelerate,yu2018nisp} including  structured~\cite{ding2019centripetal,li2016pruning} and non-structured pruning~\cite{dong2017learning,lee2019signal} is a widely used technique for reducing the computational complexity and memory usage of deep learning models by removing ``unimportant'' weights. 
Network pruning \cite{gao2021network,liang2021pruning,wang2021accelerate,yu2018nisp}, including  structured~\cite{ding2019centripetal,li2016pruning} and non-structured pruning~\cite{dong2017learning,lee2019signal}, is a widely used technique for reducing the computational complexity and memory usage of deep learning models by removing “unimportant” weights. 
However, these process often results in a significant drop in model accuracy. As the pruning ratio increases, it becomes increasingly challenging to restore the accuracy of the trimmed network. 
% To address the issue of accuracy degradation, He et al \cite{he2017channel} reconstructed the outputs with remaining channels with linear least squares and avoided the process of fine-tuning. 
To address the issue of accuracy degradation, He \textit{et al.} \cite{he2017channel} reconstructed the outputs with remaining channels with linear least squares and avoided the process of fine-tuning. 
% Zhu et al \cite{zhu2017prune} proposed an automated gradual pruning strategy in which pruning and fine-tuning are carried out alternately to prune the network gradually. 
Zhu \textit{et al.} \cite{zhu2017prune} proposed an automated gradual pruning strategy in which pruning and fine-tuning are carried out alternately to prune the network gradually. 
However, previous fine-tuning methods either bypass the fine-tuning process or aim to alleviate the accuracy drop rather than focusing on the accuracy recovery during fine-tuning. 

% \begin{figure}[t]
% \centering
% \begin{minipage}[t]{0.45\linewidth}
% \subfigure[ImageNet on ResNet-50.]{
% \label{Fig.sub.1}
% \includegraphics[width=4.4cm,height=3.3cm]{res50_point.pdf}}
% \end{minipage}
% \hfill
% \begin{minipage}[t]{0.45\linewidth}
% \subfigure[CIFAR-10 on ResNet-56.]{
% \label{Fig.sub.2}
% \includegraphics[width=4.4cm,height=3.3cm]{res56_point.pdf}}
% \end{minipage}

% \begin{minipage}[t]{0.45\linewidth}
% \subfigure[CIFAR-10 on MobileNetV2.]{
% \label{Fig.sub.3}
% \includegraphics[width=4.4cm,height=3.3cm]{mbv2_point.pdf}}
% \end{minipage}

% \caption{Performances of proposed method comparing with vanilla fine-tuning on CIFAR-10 and ImageNet.}
% \vspace{-4mm}
% \label{point_res}
% \end{figure}
% In this paper, we address this challenge by leveraging linear over-parameterization~\cite{guo2018ExpandNets,ding2021repvgg}, an approach previously used to train compact neural networks from scratch. 
% 在intro里，两次提到这篇工作解决的challenge，实际上重点只应该有一个。需要考虑我们关注的重点是“之前的方法没有真正关注fine-tune恢复精度，我们用over-p来在fine-tune恢复精度”，还是“over-p面临的问题”，确定下重点challenge（别人没有关注过没有解决的问题）后，淡化另一个（不用challenge这个词表述，可以写facing the problem of xxx, preious work tended to leverage .. method）。
% 重点应该是“之前的方法没有真正关注fine-tune恢复精度，我们用over-p来在fine-tune恢复精度”，但是over-p存在问题，使用蒸馏来解决这个问题。
In this paper, we address this challenge by leveraging linear over-parameterization~\cite{guo2018ExpandNets,ding2021repvgg}, an approach designed for training compact neural networks from scratch. 
%Linear over-parameterization was an approach to train compact neural networks from scratch. 
%It expands each linear layer of the compact network into multiple consecutive linear layers without adding any nonlinear operations and contracts it back to the original structure after training.
Linear over-parameterization generally expands linear layer of the compact network into multiple consecutive linear layers without nonlinear operations and contracts it back to the original structure after training.
% However, over-parameterization for fine-tuning after model pruning faces two challenges: 1) the loss of information due to the removal of ''unimportant'' weights. 2)  the difficulty in gathering valuable features during fine-tuning.  
However, over-parameterization for fine-tuning after model pruning faces two problems: 1) the loss of information due to the removal of “unimportant” weights, and 2) the difficulty in gathering valuable features during fine-tuning.
% new added
Facing these problems, we leverage  similarity-preserving knowledge distillation~\cite{tung2019similarity,zhang2022dense} to guide the fine-tuning of the over-parameterized network. 
% In conclusion, our approach can maintain the representation ability in fine-tuning after the expansion, and comprehensive experiments on benchmark datasets show that it outperforms existing methods, particularly for large pruning ratios. When the pruning ratio increases to an extreme 95.52\%, our method achieves 2.04\% higher accuracy restoration than vanilla fine-tuning.
% 这一句after the expansion是什么意思，maintain是和裁剪前的精度对比对吗？这里体现的意思好像是maintine了expand前的精度，这和我们要解决的challange没对应的很好。
% new added
In conclusion, our approach can better recover precision than vanilla fine-tuning, and maintain the representation ability during pruning and fine-tuning.
Comprehensive experiments show that it outperforms existing pruning methods, particularly with large pruning ratios. When the pruning ratio increases to an extreme 95.52\%, our method achieves 2.04\% higher accuracy restoration than fine-tuning itself.

\begin{figure*}[ht]
	\centering
 	\includegraphics[width=0.7\linewidth]{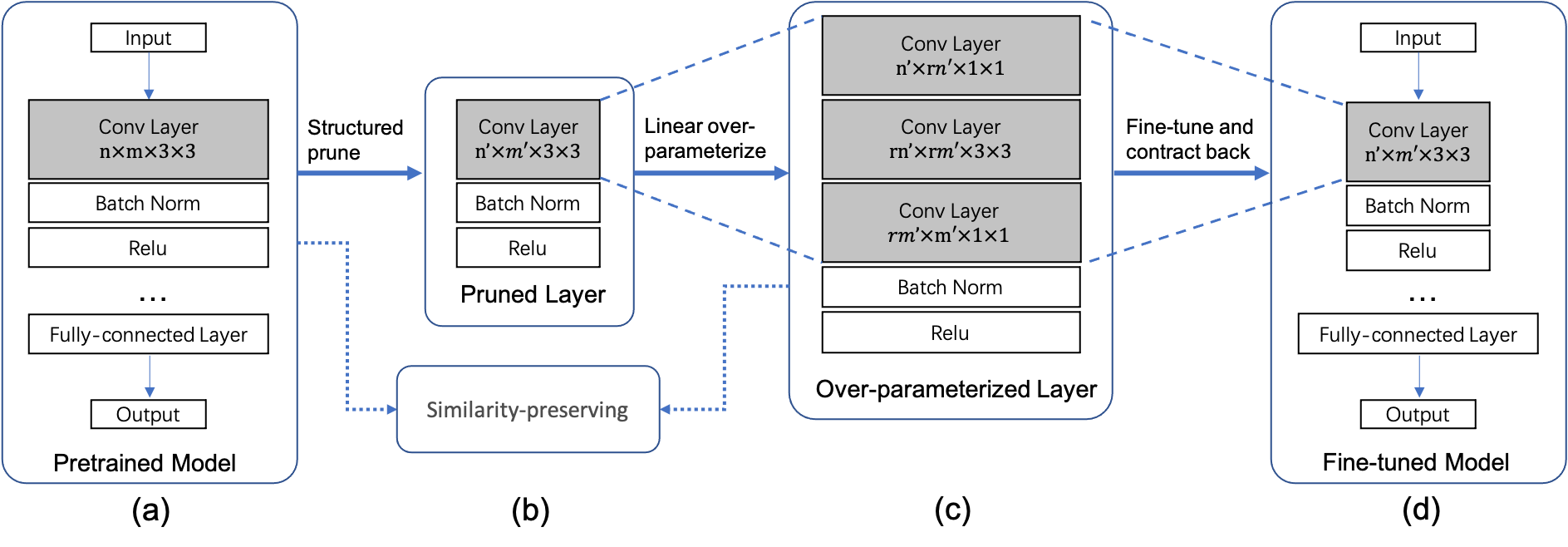}
     \caption{%Illustration of our method's overall process. (a) A dense network is obtained by pretraining on object task. (b) A pruned network is obtained by structured pruning from the pretrained network. To simplify the illustration and facilitate understanding, we just put one pruned layer instead a whole model. (c) Each pruned layer is linear over-parameterized by matrix decomposition without damage the output consistency of forward propagation before expansion. (d) After fine-tuning the over-parameterized model on object task, we contract it back to the slim model. In order to preserve the feature extraction ability of each layer after over-parameterization, pairwise activation similarities within each input mini-batch is used to supervise the training of the over-parameterized network with the pretrained network.
     An overall workflow of our method. (a) A pretrained network. (b) A pruned network. (c) Each pruned layer is linearly over-parameterized by matrix decomposition. (d) The fine-tuned over-parameterized model is contracted back to the slim model. To preserve the feature extraction ability of each layer after over-parameterization, pairwise activation similarity within each input mini-batch is used to supervise the training of the over-parameterized network.
     }
	\label{framework}
 \vspace{-4mm}
\end{figure*} 
\section{METHOD}
\label{sec:typestyle}
% In this section, we introduce our proposed approach that linearly expands the layers in the structurally pruned compact neural networks for fine-tuning. Additionally, we introduce similarity-preserving knowledge distillation for preserving feature extraction.
% 这两句称述被删的稍微有点生硬，换个方法承接上下。fig1的整体描述应该也是被删掉了？这里可以用一句话描述一下方法的整体流程
An overview of our proposed approach is illustrated in Fig.~\ref{framework}. 
%The pretrained network is structurally pruned, then linearly over-parameterized, and then fine-tuned. After fine-tuning the model is contracted back to the slim model. 
The pretrained network undergoes a sequential process of structural pruning, linear over-parameterization, fine-tuning, and contraction to achieve the final slim model.
% Additionally, in the process of fine-tuning, we introduce similarity-preserving knowledge distillation for preserving feature extraction.
During fine-tuning, similarity-preserving knowledge distillation is introduced to preserve feature extraction ability.
\subsection{Expand Layers}
\label{3.1}
% In this section, we introduce our proposed approach and
% we focus on normal convolutions, depthwise convolutions, and fully-connected layers as they are commonly used in lightweight neural networks.
% Convolutions with larger kernel size (e.g., $5\times 5$ and $7\times 7$) are not usually studied in the field of pruning since they results in more float operations per second (FLOPs) of networks.
% Therefore, the decomposition of them are out of the scope of this work although our approach can be generalized to them.
\noindent \textbf{Expanding convolutional layers.}
% We denote $\mathbf{F}_{n \times m \times k \times k}$ as the filters of a convolutional layer. 
% In this section, the filters of a convolutional layer is denoted as $\mathbf{F}_{n \times m \times k \times k}$, where $m$ denotes input channels,  $n$ denotes output channels, $k$ denotes kernel size.
In this section, the filters of a convolutional layer are represented as $\boldsymbol{F}_{n \times m \times k \times k}$, where $m$ and $n$ denote the number of input and output channels, and $k$ the kernel size.
% To expand a convolutional layer, Guo \textit{et al.} \cite{guo2018ExpandNets} proposes to expand a $k \times k$ convolutional layer into 3 consecutive convolutional layers: a $1 \times 1$ convolution; a $k \times k$ one; and another $1 \times 1$ one  (see Fig.~\ref{framework}(c)).
For a $k \times k$ convolutional layer, Guo \textit{et al.} \cite{guo2018ExpandNets} proposed to over-parameterize it into 3 consecutive layers: a $1 \times 1$ convolution; a $k \times k$ one; and another $1 \times 1$ one (Fig.~\ref{framework}(c)).
% Mathematically, for a $\mathbf{F}_{n \times m \times k \times k}$ layer, we define the number of output channels of the first $1 \times 1$ layer as $p = rm$ and the number of output channels of the intermediate $k \times k$ layer as $q = rn$ with an expansion rate $r>0$.
Based on that, for $\boldsymbol{F}_{n \times m \times k \times k}$, we define the output channel number of the first $1 \times 1$ layer as $p = rm$ and the output channel number of the intermediate $k \times k$ layer as $q = rn$ with an expansion rate $r>0$.

\begin{figure}[htbp]
\centering
\includegraphics[width=0.75\linewidth]{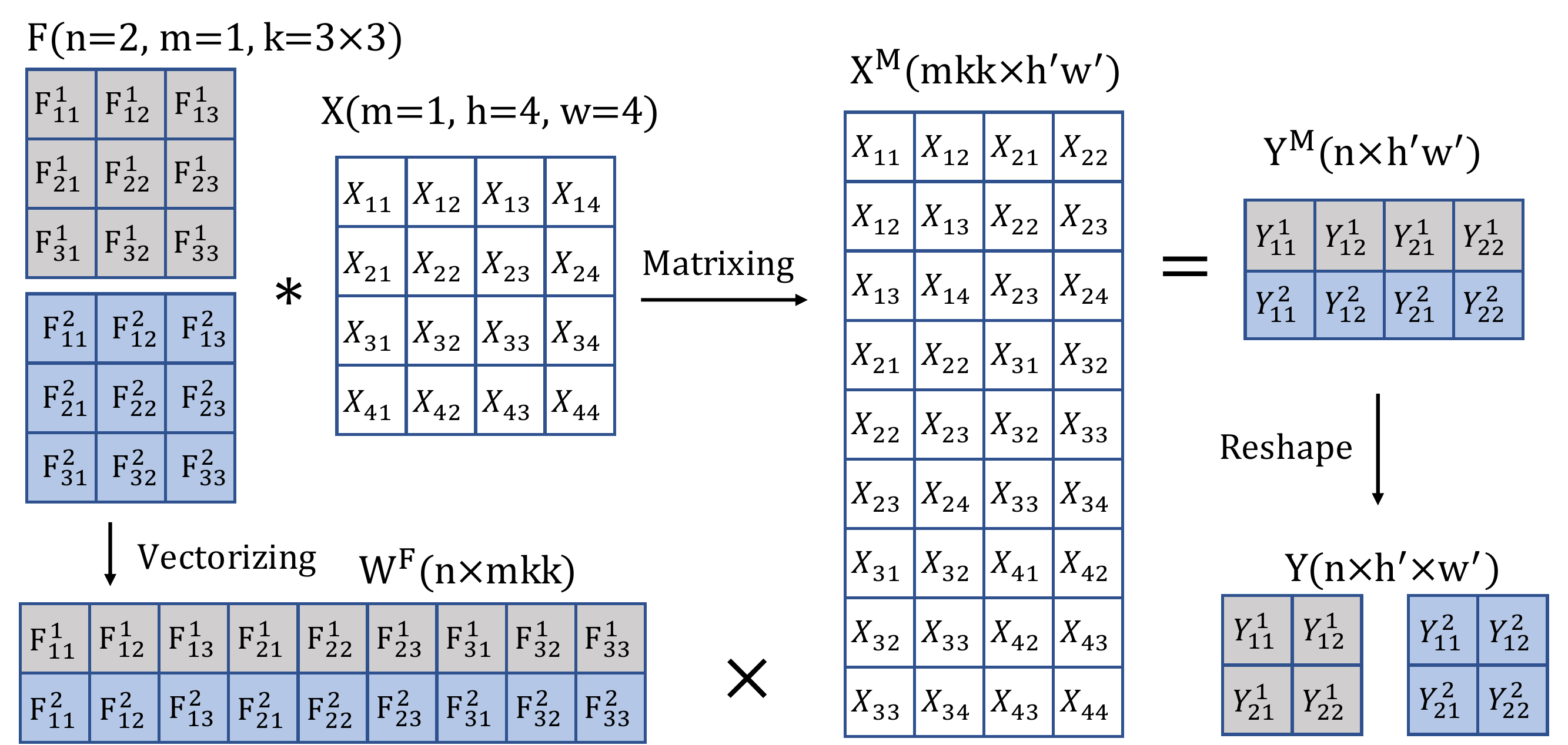}
\caption{%An illustration of the computing process of a convolutional layer. $F$ and $X$ represent the filter and input tensor, respectively.
An illustration of the over-parameterization process of a convolutional layer.
}
\label{matrix}
\vspace{-4mm}
\end{figure}

%We noticed that the convolutional filters can be expressed in matrix form, hence we can expand them using matrix decomposition, which is able to preserve and leverage the original information (weights) in pruned pretrained networks after expanding.
% Let $\mathbf{X}_{m \times h \times w}$ be the input tensor where $h$ donates the height of input tensor and $w$ donates the width of input tensor .
For a input tensor $\boldsymbol{X}_{m \times h \times w}$ with height $h$ and width $w$, the convolution can be formulated as
% Ignoring the bias for simplicity, the convolution can be expressed as
\begin{equation}\label{formula1}
\setlength{\abovedisplayskip}{3pt}
\setlength{\belowdisplayskip}{3pt}
% \begin{split}
\begin{aligned}
\setlength{\abovedisplayskip}{1pt}
\boldsymbol{Y}_{n \times h^{\prime} \times w^{\prime}}&=\boldsymbol{F}_{n \times m \times k \times k} * \boldsymbol{X}_{m \times h \times w} =\\
\operatorname{reshape}&\left(\boldsymbol{W}_{n \times mkk}^{\mathbf{F}} \times \boldsymbol{X}_{m k k \times h^{\prime} w^{\prime}}^{\mathbf{M}}\right).
\end{aligned}
% \end{split}    
\end{equation}
where $\boldsymbol{Y}_{n \times h^{\prime} \times w^{\prime}}$ is the corresponding output tensor, $h^{\prime}$ and $w^{\prime}$ are the height and width of output tensor , $\boldsymbol{X}_{m k k \times h^{\prime} w^{\prime}}^{\mathbf{M}}$ is the unrolled matrix representation of the input tensor $\boldsymbol{X}_{m \times h \times w}$, and $\boldsymbol{W}_{n \times mkk}^{\boldsymbol{F}}$ is the matrix representation of the convolutional filters $\boldsymbol{F}_{n \times m \times k \times k}$. 
% Fig.~\ref{matrix} shows an example of this computational process.
Fig.~\ref{matrix} provides an illustration of this process.
% With this matrix representation, we can therefore expand a layer linearly by replacing the matrix with the other three matrices. 
Based on formula  \ref{formula1}, a layer can be expanded linearly by replacing one matrix with matrices multiplied in succession. 

% To be more specific, we can express the form of matrix multiplication
More specifically, the matrix multiplication is formulated as 
\begin{equation}\label{formula2}
\setlength{\abovedisplayskip}{3pt}
\setlength{\belowdisplayskip}{3pt}
\begin{aligned}
% \boldsymbol{W}_{mkk\times n}^{\mathbf{F}} = \operatorname{reshape}(\boldsymbol{W}_{m\times p}^{\mathbf{F}^1} \times \boldsymbol{W}_{p\times kkq}^{\mathbf{F}^{2}})_{mkk\times q}\times \boldsymbol{W}_{q\times n}^{\mathbf{F}^3}, \\
&\boldsymbol{W}_{n \times mkk}^{\mathbf{F}} = \boldsymbol{W}_{n\times q}^{\mathbf{F}^3}  \\&\times \operatorname{reshape}(\operatorname{reshape}(\boldsymbol{W}_{q\times pkk}^{\mathbf{F}^{2}})_{qkk\times p} \times \boldsymbol{W}_{p\times m}^{\mathbf{F}^1})_{q\times mkk}.
\end{aligned}
\end{equation}
% Then we can expand the layers by solving and assigning these three matrices. 
Based on formula \ref{formula2}, the expanded layers are determined by matrix solving.
% We first initializing $\boldsymbol{W}_{n\times q}^{\mathbf{F}^3}$ and $\boldsymbol{W}_{p\times m}^{\mathbf{F}^1}$ randomly and then calculate the weight of the rest layer by using matrix decomposition. 
More specifically, $\boldsymbol{W}_{n\times q}^{\mathbf{F}^3}$ and $\boldsymbol{W}_{p\times m}^{\mathbf{F}^1}$ are randomly initialized, so the weight of $\boldsymbol{W}_{q\times pkk}^{\mathbf{F}^{2}}$ can be determined by matrix decomposition.
This process can be formulated as
\begin{equation}
\begin{aligned}
&\boldsymbol{W}_{q\times pkk}^{\mathbf{F}^2} =(\boldsymbol{W}_{n\times q}^{\mathbf{F}^3})^{-1}_R \\ &\times \operatorname{reshape}(\operatorname{reshape}(\boldsymbol{W}_{n\times mkk}^{\mathbf{F}})_{nkk\times m}\times (\boldsymbol{W}_{p\times m}^{\mathbf{F}^1})_L^{-1})_{n\times pkk}.
% \boldsymbol{W}_{p\times kkq}^{\mathbf{F}^2} &=(\boldsymbol{W}^{\mathbf{F}^1})^{-1}_{p\times m}\\ &\times\operatorname{reshape}( \boldsymbol{W}_{mkk\times n}^{\mathbf{F}}\times (\boldsymbol{W}^{\mathbf{F}^3})_{n\times q}^{-1})_{m\times kkq},
\label{eq_expand_conv}
    \end{aligned}
\end{equation}
% where $(\boldsymbol{W}_{n\times q}^{\mathbf{F}^3})^{-1}_R$ indicates the right inverse of $\boldsymbol{W}_{n\times q}^{\mathbf{F}^3}$. $(\boldsymbol{W}_{p\times m}^{\mathbf{F}^1})_L^{-1}$ means the left inverse of $\boldsymbol{W}_{p\times m}^{\mathbf{F}^1}$ .
where $(\boldsymbol{W}_{n\times q}^{\mathbf{F}^3})^{-1}_R$ indicates the right inverse of $\boldsymbol{W}_{n\times q}^{\mathbf{F}^3}$ and $(\boldsymbol{W}_{p\times m}^{\mathbf{F}^1})_L^{-1}$, the left inverse of $\boldsymbol{W}_{p\times m}^{\mathbf{F}^1}$ .
% In this way, we can expand the convolution structure and reshape it to the original compact network algebraically after fine-tuning (see Fig.~\ref{framework}(d)).
After similarity-preserving guided fine-tuning, the decomposed parameters go through matrix reconstruction following the decomposition pathway into the compact shape (Fig.~\ref{framework}(d)).
%After hard-pruning the network, we have obtained the weight $\boldsymbol{W}_{m k k \times n}^{\mathbf{F}}$ determined by each layer. 
%Then we assign the weights of the three layers by randomly initializing two of them and calculating the weight of the rest layer by using matrix decomposition.
To expand a convolutional layer with padding $p$, we apply padding $p$ in the first layer of the expanded unit while not padding the remaining layers. To handle a stride $s$, we set the stride of the middle layer to $s$ and that of the others to 1.

\noindent \textbf{Expanding depthwise convolutional layers.} 
% Depthwise convolutions are often used to design compact networks, such as MobileNetV2~\cite{sandler2018mobilenetv2} and ShuffleNetV2~\cite{ma2018shufflenet}. 
% %To expand the depthwise convolutional layers, we set the group of the three consecutive linear expand layers same as the original layer, then we make use of our normal convolutional expansion strategy within each group. This makes the expanded layers equivalent to the normal ones for each group.
To expand the depthwise convolutional layers, we set the group parameter of the expanded layers same as the original layer, then the normal convolutional expansion strategy is applied on these three consecutive layers within each group. This makes the expanded layers equivalent to the original ones for each group.

\noindent \textbf{Expanding fully-connected layers.} 
%Here we introduce the approach of expanding a fully-connected layer to a series of fully connected layers. 
As for a  fully-connected layer, we can directly expand a linear layer with $m$ input and $n$ output dimensions into $l$ linear layers as
\begin{equation}\label{eq5}
\boldsymbol{W}_{n \times m}=\boldsymbol{W}_{n \times p_{l-1}} \times \boldsymbol{W}_{p_{l-1} \times p_{l-2}} \times \cdots \times \boldsymbol{W}_{p_{1} \times m}.
\end{equation}
%Through experiments, compared with \cite{arora2018optimization}, expanding only fully-connected layers does typically not yield a performance boost, \cite{guo2018ExpandNets} proved that expand both convolutional layers and fully-connected layers simultaneously can bring performance improvement. 
We allocate the weights of the expanding layers by randomly initializing $l-1$ of them and calculating the weight of the remaining $\theta$-th layer by matrix decomposition.
\begin{equation}
\setlength{\abovedisplayskip}{3pt}
\setlength{\belowdisplayskip}{3pt}
\begin{aligned}
    \boldsymbol{W}_{p_{\theta} \times p_{\theta-1}} &=\prod_{i=\theta-1}^{l}\left(\boldsymbol{W}_{p_{i} \times p_{i-1}}\right)_{R}^{-1}  \\&\times \boldsymbol{W}_{n \times m} \times \prod_{i=1}^{\theta-1}\left(\boldsymbol{W}_{p_{i} \times p_{i-1}}\right)_{L}^{-1}.
    \label{eq_linear_expand}
    \end{aligned}  
\end{equation}

For convenience, we denote $p_{l} = n$ and $p_{0} = m$ in Eq.~\ref{eq5} to get Eq.~\ref{eq_linear_expand}. In practice, considering the computational complexity of fully-connected layers, we expand each layer into only two or three layers with a small expansion rate. 

\noindent \textbf{Existence of left and right inverses.} 
%bn ,padding, stride, inverse
% Normal convolutional layer and fully-connected layer are often followed by batch normalization (BN). When we expand these layers, it is technically feasible for us to add BN after each expand layer. However, BN layers contain statistical information on the training set. And these information will lost through initializing convolution and full-connected layers then calculating the corresponding BN. Therefore, we do not add BN to the expand layers and just maintain the BN followed to the original layer. 
%The left and right inverses of matrices need to be computed in Eq.~\ref{eq_expand_conv} and Eq.~\ref{eq_linear_expand}. Therefore, the left and right inverses of these matrices must be determined to exist (if not, we can only compute the pseudo inverse of the matrix. Then the difference will be caused between the expand layers and the original layer).
The left and right inverses of matrices need to be computed in Eq.~\ref{eq_expand_conv} and Eq.~\ref{eq_linear_expand} so they must exist. 
%We considering the matrix $\boldsymbol{W}_{n \times m}$, the necessary and sufficient condition for the existence of its right inverse is $n \leq m$ and $rank(\boldsymbol{W}_{n \times m}) = n$ , i.e., matrix $\boldsymbol{W}_{n \times m}$ is row full rank. In our work, all matrices requiring inversion are initialized randomly. When $\boldsymbol{W}_{n \times m}$ is generated by random with normal distribution, we can get $\boldsymbol{W}_{n \times n}^{'}$ after we removed the last $m-n$ columns from $\boldsymbol{W}_{n \times m}$. Then we can easily get:
In our work, all matrices that need determining the inverses are initialized randomly. Considering matrix $\boldsymbol{W}_{n \times m}$, the necessary and sufficient condition for the existence of its right inverse is $n \leq m$ and $rank(\boldsymbol{W}_{n \times m}) = n$ , i.e., matrix $\boldsymbol{W}_{n \times m}$ is row full rank. As $\boldsymbol{W}_{n \times m}$ is randomly initialized with normal distribution, the inverse $\boldsymbol{W}_{n \times n}^{'}$ can be determined after we remove the last $m-n$ columns. Under these circumstances, we can compare their full rank probability as
\begin{equation}\label{eq7}
P\left[\operatorname{rank}\left(\boldsymbol{W}_{n \times m}\right)=n\right] \geq P\left[\operatorname{rank}\left(\boldsymbol{W}_{n \times n}^{\prime}\right)=n\right].
\end{equation}
%\cite{feng2007rank} proved that a rational random matrix has full rank with probability 1, that is, $P\left[\operatorname{rank}\left(\boldsymbol{W}_{n \times n}^{\prime}\right)=n\right]=1$. Then we can get $P\left[\operatorname{rank}\left(\boldsymbol{W}_{n \times m}\right)=n\right]=1$, so its right inverse exit certainly. So the idea to ensure the existence of a matrix's left and right inverse is to control the expansion rate. E.g., if we need to expand a fully-connected layer $\boldsymbol{W}_{n \times m}$ to $\boldsymbol{W}_{n \times p} \times \boldsymbol{W}_{p \times m}$. If we randomly initialize $\boldsymbol{W}_{n \times p}$ and calculate $\boldsymbol{W}_{p \times m}$, we should set the value $p \geq n$. Conversely, we should set the value $p \geq m$.  
Feng \textit{et al.} \cite{feng2007rank} has proved that a rational random matrix has full rank with probability 1, that is, $P\left[\operatorname{rank}\left(\boldsymbol{W}_{n \times n}^{\prime}\right)=
n\right]=1$. Thus with Eq.~\ref{eq7} we get $P\left[\operatorname{rank}\left(\boldsymbol{W}_{n \times m}\right)=n\right]=1$ and its right inverse exist certainly. In this way, in order to ensure the existence of a matrix's left and right inverse, what we need to do is to control the expansion rate. E.g., when we expand a fully-connected layer $\boldsymbol{W}_{n \times m}$ to $\boldsymbol{W}_{n \times p} \times \boldsymbol{W}_{p \times m}$, if we randomly initialize $\boldsymbol{W}_{n \times p}$ and calculate $\boldsymbol{W}_{p \times m}$, we should ensure $p \geq n$. Conversely, we should set the value $p \geq m$.  
\subsection{Feature Extraction Preserving} 

% After expanding the layers, the expanded weights are constructed based on random initialization and matrix decomposition which can  not hold a stable feature extraction ability as the original network. To handle this, similarity-preserving knowledge distillation is applied. 
%It guides the fine-tuning of over-parameterized networks such that produce similar activations in the un-pruned pretrained network produce similar activations in the over-parameterized pruned network. 

Over-parameterization will destroy the stable feature extraction ability of the pretrained and pruned model. To handle this, similarity-preserving knowledge distillation~\cite{tung2019similarity} is applied. Given an input mini-batch, it denotes the activation map produced by the un-pruned network T (teacher network) at a particular layer $l$ as $(A_{T}^{(l)})_{b \times m \times h \times w}$ and the activation map produced by the over-parameterized network S (student network) at the last corresponding expand layer $l^{\prime}$ as $(A_{S}^{(l^{\prime})})_{b \times m^{\prime} \times h^{\prime} \times w^{\prime}}$. 
The distillation loss, an L2 regularization penalty, is applied. First, let $(Q_{T}^{(l)})_{b \times mhw} = {reshape((A_{T}^{(l)})_{b \times m \times h \times w})}_{b \times mhw}$. Then we get
\begin{equation}
(\tilde{G}_{T}^{(l)})_{b \times b}=(Q_{T}^{(l)})_{b \times mhw} \cdot (Q_{T}^{(l) \top })_{mhw \times b},
\end{equation}
\begin{equation}
G_{T[i,:]}^{(l)}=\tilde{G}_{T[i,:]}^{(l)} /\left\|\tilde{G}_{T[i,:]]}^{(l)}\right\|_{2}.
\end{equation}
\setlength{\abovedisplayskip}{3pt}
\setlength{\belowdisplayskip}{3pt}
\begin{equation}
(\tilde{G}_{S}^{(l^{\prime})})_{b \times b}=(Q_{S}^{(l^{\prime})})_{b \times m^{\prime}h^{\prime}w^{\prime}} \cdot (Q_{S}^{(l^{\prime}) \top })_{m^{\prime}h^{\prime}w^{\prime} \times b},
\end{equation}
\begin{equation}
G_{S[i,:]}^{(l^{\prime})}=\tilde{G}_{S[i,:]}^{(l^{\prime})} /\left\|\tilde{G}_{S[i,:]]}^{(l^{\prime})}\right\|_{2}.
\end{equation}
The similarity-preserving knowledge distillation loss is designed as
\begin{equation}
\mathcal{L}_{\mathrm{SP}}\left(G_{T}, G_{S}\right)=\frac{1}{b^{2}} \sum_{\left(l, l^{\prime}\right) \in \mathcal{I}}\left\|G_{T}^{(l)}-G_{S}^{\left(l^{\prime}\right)}\right\|_{F}^{2}.
\label{eq_sp}
\end{equation}
%where $\mathcal{I}$ collects the $\left(l, l^{\prime}\right)$ layer pairs (e.g. layers in the un-pruned network and their last corresponding expand layers in the over-parameterized network.) and $\|\cdot\|_{F}$ is the Frobenius norm. Eq.~\ref{eq_sp} is a summation, over all $\left(l, l^{\prime}\right)$ pairs, of the mean element-wise squared difference between the $G_{T}^{(l)}$ and $G_{S}^{\left(l^{\prime}\right)}$ matrices. Finally, we define the total loss for training the student network as:
where $\mathcal{I}$ collects the $\left(l, l^{\prime}\right)$ layer pairs. Finally, the total loss for student network training is defined as
\begin{equation}
\mathcal{L}=\mathcal{L}_{\mathrm{task}}\left(\mathbf{Y}, \sigma_{\mathrm{S}}\left(\mathbf{X}\right)\right)+\gamma \mathcal{L}_{\mathrm{SP}}\left(G_{T}, G_{S}\right).
\end{equation}
where $\mathcal{L}_{\mathrm{task}}(\cdot, \cdot)$ denotes the loss function of vision task, $\mathbf{Y}$ is the ground truth, and $\mathbf{X}$ is the input mini-batch. $\sigma_{\mathrm{S}}(\cdot)$ denotes the output from over-parameterized network S.

\section{EXPERIMENTS}
\label{sec:majhead}

\subsection{Implemented Details}
%The input image size is 224$\times$224, and the mini-batch size is 256 trained on 8 GPUs. The learning rate is initialized to 0.03 and subjects to cosine learning rate schedule. Most of the experiments are conducted on the CIFAR-10 dataset, though we also report some results on CIFAR-100. When the pruned model is transferred to the downstream task, a supervised linear classifier (usually a fully-connected layer) is trained with the pruned model fixed. Then the retrained model is evaluated with top-1 classification accuracy. For all experiments, the MoCo is pretrained for 1000 epochs, and then the linear classifier is retrained for 300 epochs to get the final results.
%The model is trained on 8 GPUs with the input image size 224$\times$224 and the mini-batch size 256. The learning rate is initialized to 0.03 and subjects to cosine learning rate schedule. Most experiments are conducted on the CIFAR-10 dataset, though we also report some results on CIFAR-100. When the pruned model is transferred to the downstream task, a supervised linear classifier (usually a fully-connected layer) is trained with the pruned model fixed. Then the retrained model is evaluated with top-1 classification accuracy. For all experiments, the model is pretrained for 1000 epochs, and then the linear classifier is retrained for 300 epochs to get the final results.

In our experiments, we evaluate the proposed strategies on CIFAR-10~\cite{krizhevsky2009learning} and ImageNet~\cite{deng2009imagenet}.
The image size is 32$\times$32 for CIFAR-10 and 224$\times$224 for ImageNet, and  trained with the batch size 256. We apply the ADMM strategy and compare the effect of our proposed method and other fine-tuning methods in the fine-tuning stage. To demonstrate the generalization of our proposed method, we conduct comparative experiments under other pruning strategies, including AMC \cite{he2018amc}, HRank \cite{lin2020hrank}, ThiNet \cite{luo2017thinet}, and EagleEye \cite{li2020eagleeye}. 
% To prove the effectiveness of our proposed method, we apply the ADMM strategy for compression and then compare the effect of our proposed method and other fine-tuning methods in the fine-tuning stage. Because ADMM is a commonly used pruning strategy in the industry, it is representative to verify our method on this method.
% Moreover, to demonstrate the generality of our proposed method, we conduct comparative experiments under more pruning methods, like AMC \cite{he2018amc}, HRank \cite{lin2020hrank}, ThiNet \cite{luo2017thinet}, and EagleEye \cite{li2020eagleeye}.

\subsection{Results}
%In this section, our algorithm is compared with four other methods. Firstly, after pruning the network with a reset pruning ratio from the pretrained model, these five retrain methods can all guarantee that the final model used for inference has the same structure with the pruned model, that is, they have the same FLOPSs and latency. We compared these five methods under various proportion of removed weight. We should notice that after removing the filters of a layer for structured pruning, the input channel of the next layer will be removed passively, so the ultimate percent of removed weights are always larger than our preset prune ratio and presented as irregular numbers. The other four methods are: 1) reinitialize the pruned model and retrain it from scratch. 2) vanilla fine-tuning. 3) expand the pruned model by initializing the weights and contract it back after training. 4) reinitialize the pruned model and retrain it guided by the similarity-preserving knowledge distillation loss from the pretrained model. In Sec.~\ref{as}, we performed ablation study on expansion rate and set it to 3 is enough through observation. So we set the expansion rate in our algorithm and ExpandNets to 3 in the following experiments. 

\subsubsection{Results for ADMM}
% There are two cases in pruned network: 1) utilize its weights during fine-tuning 2) keep only the structure of the pruned network. 
After removing the less important parameters, there are two styles of fine-tuning: (1) retaining the remaining weights and (2) training the pruned structure from scratch. 
For the first case, we fine-tune the network for 150 epochs, with a learning rate of 0.01 decay by a factor 10 at epochs 60, 100, and 120. We retrain the network for 280 epochs for the second case, with a learning rate of 0.1 decayed by a factor 10 at epochs 100, 160, 210, and 250.

In this section, we compare the proposed method with four other competing methods. These methods include: 1) training from scratch, 2) ExpandNets~\cite{guo2018ExpandNets}, 3) vanilla distillation, where we retrain the reinitialized model with similarity-preserving knowledge distillation loss, and 4) vanilla fine-tuning.
\begin{table}[htbp]
  \centering
  \setlength{\tabcolsep}{0.9mm}{
    \begin{tabular}{cccccccccc}
    \toprule
    \multirow{2}{*}{\tabincell{c}{\textbf{ADMM}}} &
    \multicolumn{9}{c}{Percent of weights removed (\%)}\cr
    \cmidrule(lr){2-10}
                                &0	&68.37		&78.23				&94.80	&98.66\cr
    \midrule
    Training from scratch&94.17	&86.22		&83.47				&72.08	&49.08\cr
    ExpandNets&\textbf{94.36}	&87.61		&84.99				&75.14	&51.25\cr
    Vanilla distillation&94.17	&87.80		&85.03				&75.03	&51.31\cr
    Vanilla fine-tuning&94.17	&93.27		&92.74				&86.03	&54.39\cr
    \textbf{Ours}&94.17	&\textbf{93.31}		&\textbf{93.26}			&\textbf{87.97}	&\textbf{56.88}\cr

    \bottomrule
    \end{tabular}
    \caption{Top-1 accuracy(\%) on the test set of CIFAR-10 based on MobileNetV2.}
    \label{tbl:mbv2_cifar10}
    }
    \vspace{-2mm}
\end{table}

\noindent \textbf{MobileNetV2 on CIFAR-10.} 
Tab.~\ref{tbl:mbv2_cifar10} presents the results for all  methods. Our method outperforms the other methods in terms of accuracy restoration, especially when the pruning ratio is relatively large. ExpandNets and vanilla distillation produce similar results, which is consistent with~\cite{guo2018ExpandNets}. Both methods perform much worse than vanilla fine-tuning, indicating that applying over-parameterization and knowledge distillation alone is not sufficient for pruning.

\noindent \textbf{ResNet-50 on ImageNet.} 
% For ResNet-50 on ImageNet, the top-1 accuracy results in Tab.~\ref{tbl:rs50_img} are consistent with the conclusions above. Notably, our method does not perform well with a small pruning ratio, and it is even worse than vanilla fine-tuning when the percentage of removed weights is 28.37\%. Moreover, when the network is not pruned, the performance of ExpandNets is not better than that of the direct pretrained model. We believe that the reason for this phenomenon is that when the parameters are sufficient, over-parameterization will only make the network training more cumbersome and slow down the convergence.
For ResNet-50 on ImageNet, the top-1 accuracy results are shown in Tab.~\ref{tbl:rs50_img}, which are consistent with the conclusions above.
\begin{table}[ht]\small
  \centering
  \setlength{\tabcolsep}{0.8mm}{
    \begin{tabular}{cccccc}
    \toprule
    \multirow{2}{*}{\tabincell{c}{\textbf{ADMM}}} &
    \multicolumn{5}{c}{Percent of weights removed (\%)}\cr
    \cmidrule(lr){2-6}
    &0	&28.37	&54.25	&71.23	&93.18\cr
    \midrule
    Training from scratch&79.18	&74.14	&71.42	&67.39	&56.79\cr
    ExpandNets&78.87	&74.28	&73.82	&71.06	&58.07\cr
    Vanilla distillation&79.18	&75.32	&73.69	&72.61	&58.65\cr
    Vanilla fine-tuning&79.18	&\textbf{77.63}	&75.03	&72.88	&62.47\cr
    \textbf{Ours}&\textbf{79.18}	&77.47	&\textbf{75.76}	&\textbf{73.36}	&\textbf{64.08}\cr

    \bottomrule
    \end{tabular}
    \caption{Top-1 accuracy(\%) on the test set of ImageNet based on ResNet-50.}
    \label{tbl:rs50_img}
    }
    \vspace{-4mm}
\end{table}
 \subsubsection{Results for other pruning strategies}
\begin{figure}[htbp]
\centering  %图片全局居中
\subfigure[HRank]{
\label{Fig.sub.21}\hspace{-4mm}
\includegraphics[width=2.9cm,height = 2.9cm]{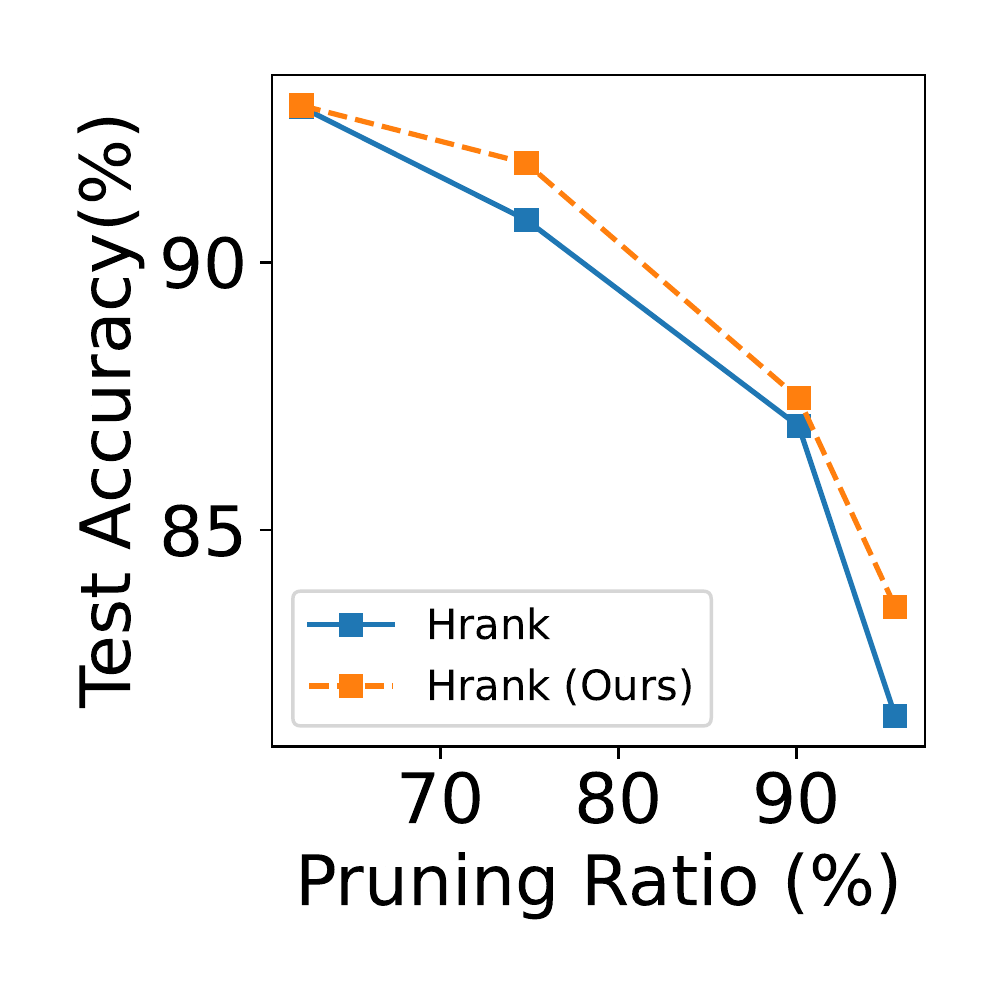}}
\subfigure[ThiNet]{
\label{Fig.sub.31}\hspace{-4mm}
\includegraphics[width=2.9cm,height = 2.9cm]{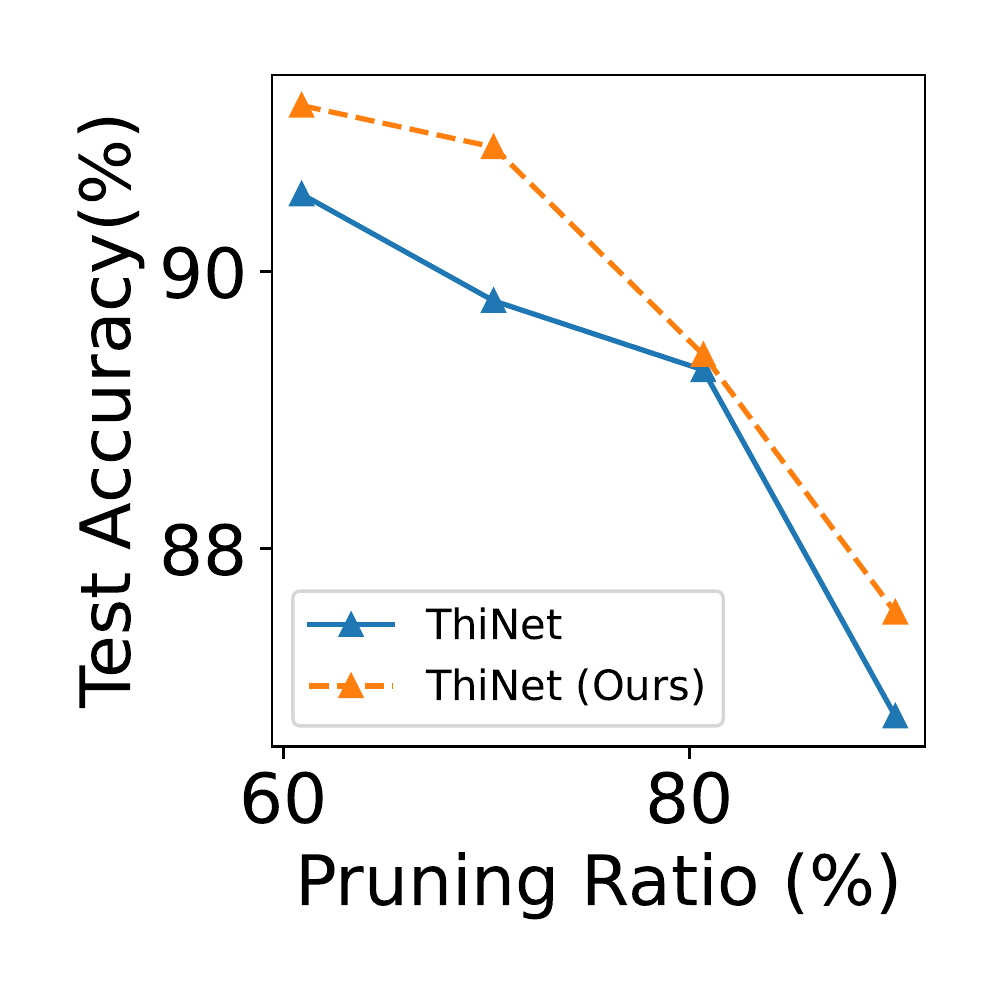}}
\subfigure[Eagleeye]{
\label{Fig.sub.41}\hspace{-4mm}
\includegraphics[width=2.9cm,height = 2.9cm]{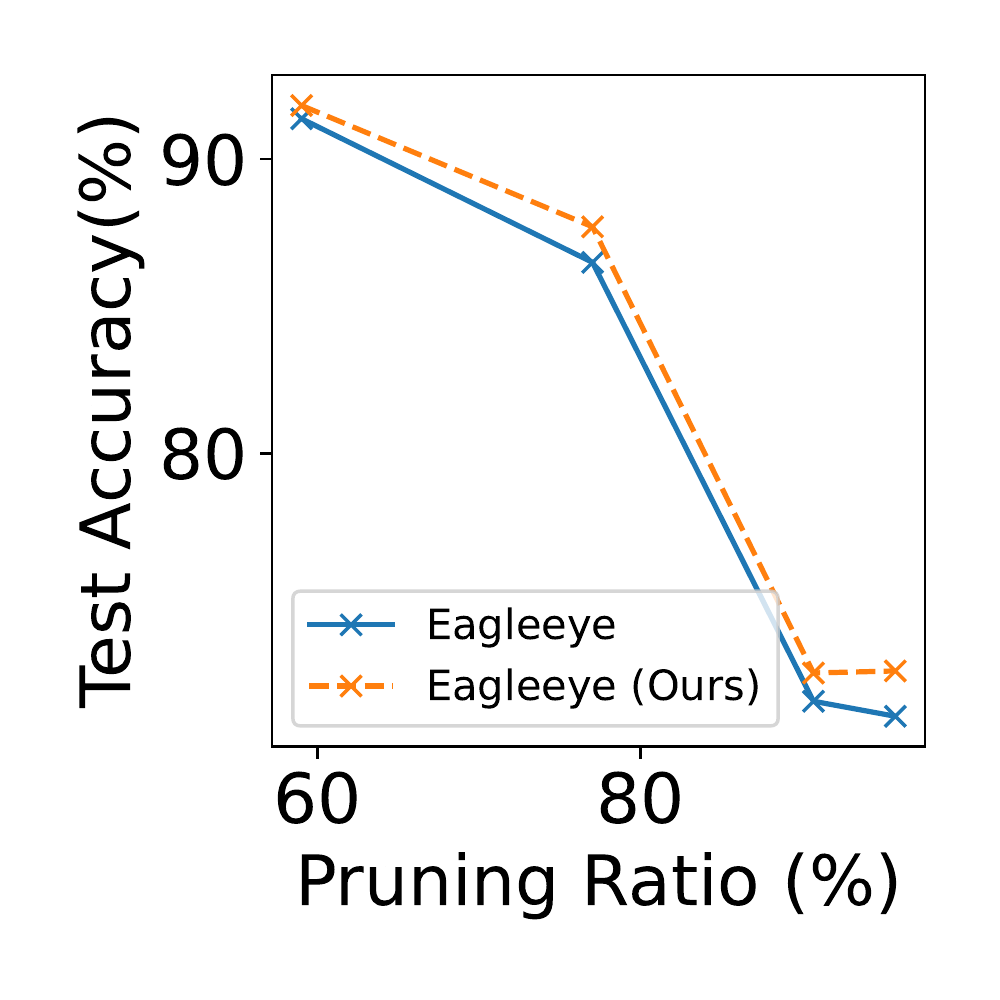}}\hspace{-4mm}
\caption{Top-1 accuracy(\%) on the test set of CIFAR-10 based on ResNet-56.}
\label{res56_liner}
\vspace{-5mm}
\end{figure}

\begin{figure}[htbp]
\centering  %图片全局居中
\subfigure[AMC]{
\label{Fig.sub.111}\hspace{-4mm}
\includegraphics[width=2.9cm,height = 2.9cm]{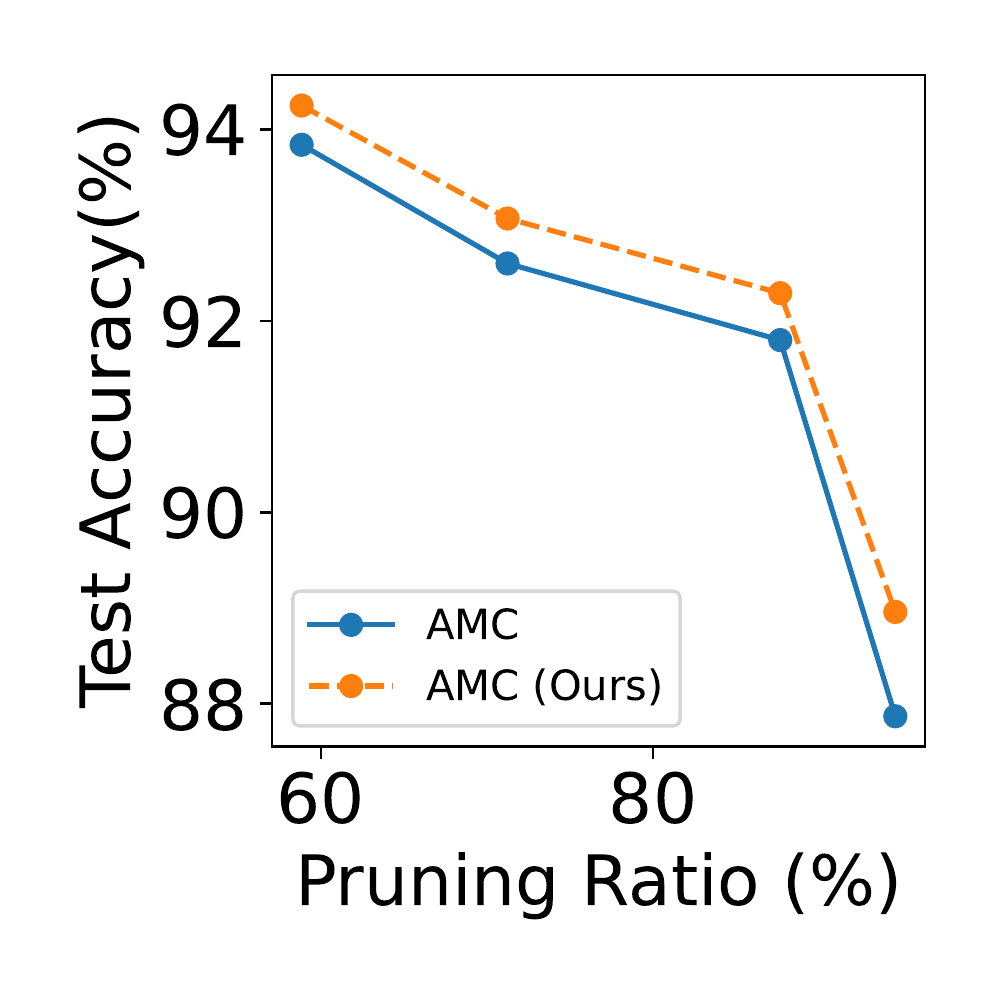}}
\subfigure[HRank]{
\label{Fig.sub.211}\hspace{-4mm}
\includegraphics[width=2.9cm,height = 2.9cm]{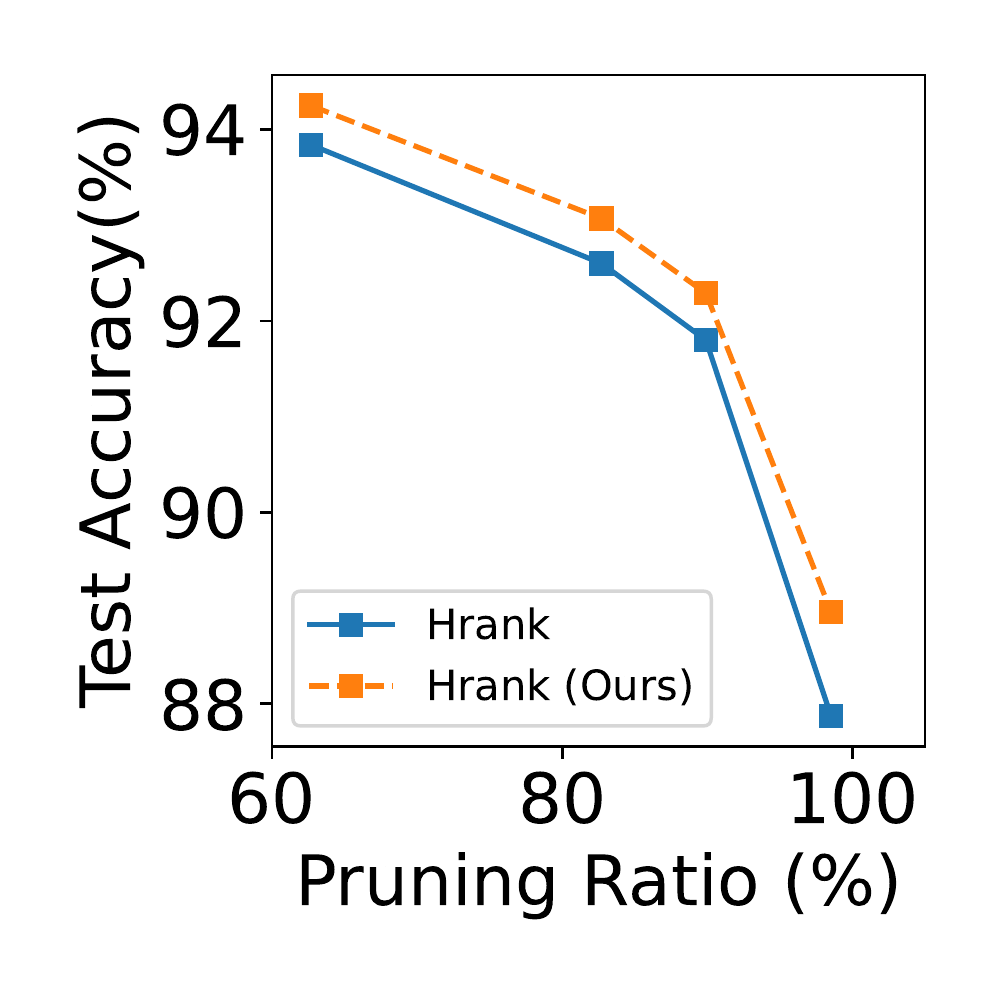}}
\subfigure[Eagleeye]{
\label{Fig.sub.411}\hspace{-4mm}
\includegraphics[width=2.9cm,height = 2.9cm]{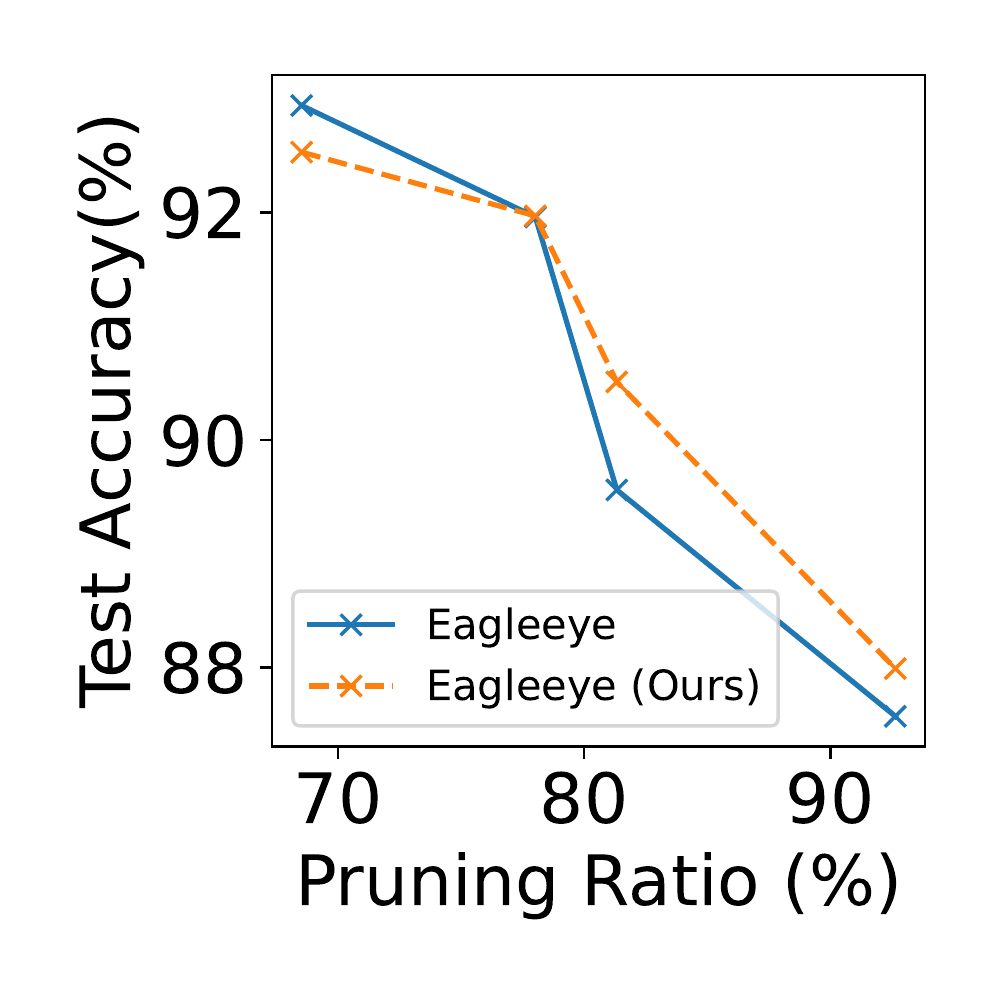}}\hspace{-4mm}
\caption{Top-1 accuracy(\%) on the test set of CIFAR-10 based on MobileNetV2.}
\label{mbv2_liner}
\vspace{-5mm}
\end{figure}

In this section, we apply various pruning strategies to prove the reliability of the proposed method. And we compare the proposed with  Vanilla fine-tuning.
%\footnote{For the Vanilla fine-tuning on CIFAR-10 and  ImageNet, We train the network with 100 epochs and adjust the learning rate using CosineAnnea. For the proposed, we train the network with 200 epochs and adjust the learning rate using CosineAnnea on CIFIAR-10. And train the network with 150 epochs decaying the ratio by a factor 10 at epochs 60,100,120 on ImageNet. }.
Fig.~\ref{res56_liner}, Fig.~\ref{mbv2_liner} and Tab.~\ref{tbl:res50} show that in most cases, our proposed method is superior to Vanilla fine-tuning.

% \begin{figure}[htb]
% \centering  %图片全局居中
% % \setlength{\abovecaptionskip}{0.cm}
% % \subfigure[ADMM]{
% % \label{Fig.sub.11}
% % \includegraphics[width=2.9cm,height = 2.9cm]{liner1_1.pdf}}
% \subfigure[HRank]{
% \label{Fig.sub.21}\hspace{-4mm}
% \includegraphics[width=2.9cm,height = 2.9cm]{liner1_2.pdf}}
% \subfigure[ThiNet]{
% \label{Fig.sub.31}\hspace{-4mm}
% \includegraphics[width=2.9cm,height = 2.9cm]{liner1_3.pdf}}
% \subfigure[Eagleeye]{
% \label{Fig.sub.41}\hspace{-4mm}
% \includegraphics[width=2.9cm,height = 2.9cm]{liner1_4.pdf}}\hspace{-4mm}
% \caption{Top-1 accuracy(\%) on the test set of CIFAR-10 based on ResNet-56.}
% \label{res56_liner}
% \vspace{-5mm}
% \end{figure}

% \begin{figure}[htb]
% \centering  %图片全局居中
% \subfigure[AMC]{
% \label{Fig.sub.111}\hspace{-4mm}
% \includegraphics[width=2.9cm,height = 2.9cm]{liner2_1.pdf}}
% \subfigure[HRank]{
% \label{Fig.sub.211}\hspace{-4mm}
% \includegraphics[width=2.9cm,height = 2.9cm]{liner2_2.pdf}}
% % \subfigure[ThiNet]{
% % \label{Fig.sub.311}\hspace{-4mm}
% % \includegraphics[width=2.9cm,height = 2.9cm]{liner2_3.pdf}}
% \subfigure[Eagleeye]{
% \label{Fig.sub.411}\hspace{-4mm}
% \includegraphics[width=2.9cm,height = 2.9cm]{liner2_4.pdf}}
% \caption{Top-1 accuracy(\%) on the test set of CIFAR-10 based on MobileNetV2.}
% \label{mbv2_liner}
% \vspace{-5mm}
% \end{figure}

% \begin{figure*}[ht]
% 	\centering
%  	\includegraphics[width=0.9\linewidth]{mbv2_liner.jpg }
%      \caption{Top-1 accuracy(\%) on the test set of CIFAR-10 based on MobileNetV2 applying other strategies.
%      }
% 	\label{mbv2_liner}
%  \end{figure*}

 \begin{table}[H]
  \centering
  \setlength{\tabcolsep}{2mm}{
    \begin{tabular}{ccccc}
    \toprule
    % AMC
    \multirow{2}{*}{\tabincell{c}{\textbf{AMC}}} &
    \multicolumn{4}{c}{Percent of weights removed (\%)}\cr
    \cmidrule(lr){2-5}
    &0&52.64 &63.28 &95.63 \cr    
    \midrule
    Vanilla fine-tuning &80.10 &74.60 &72.30 & 50.52	\cr
    \textbf{Ours} &80.10&\textbf{75.50} & \textbf{73.15} &\textbf{51.20}\cr
    
    \bottomrule
    \end{tabular}
    \caption{Top-1 accuracy(\%) on the test set of ImageNet based on ResNet-50.}
    \label{tbl:res50}
    }
\end{table}

\subsection{Ablation Study}\label{as}
\noindent \textbf{Similarity-preserving knowledge distillation.} %We do the ablation experiments on CIFAR-10 based on MobileNetV2. The results are shown in Tab..~\ref{tbl:ab_distall}. The results by over-parameterized fine-tuning with out knowledge distillation performs slightly higher than vanilla fine-tuning but unstable. When we removed 86.2\% weights, its top-1 accuracy can only achieve 92.24\% which is 0.06\% lower than vanilla fine-tuning. We think the reason is that the expanded weights are constructed based on randomly initialization and matrix decomposition, therefore they do not hold a stable feature extraction ability as the original network. When we used the similarity-preserving knowledge distillation to guide the over-parameterized network fine-tuning, our method can achieve better and more stable performance.
The ablation study is carried out on CIFAR-10 with MobileNetV2. The results are shown in Tab.~\ref{tbl:ab_distall}. Over-parameterized fine-tuning without knowledge distillation performs slightly better than vanilla fine-tuning but is unstable. When we use similarity-preserving knowledge distillation, our method can achieve better and more stable performance.
\begin{table}[htbp]
  \centering
    \setlength{\tabcolsep}{0.3mm}{\begin{tabular}{ccccccccccc}
    \toprule
    \multirow{2}{*}{\setlength{\tabcolsep}{0.2mm}\tabincell{c}{\textbf{ADMM}}} &
    \multirow{2}{*}{\setlength{\tabcolsep}{0.2mm}\tabincell{c}{Distillation}} &
    \multicolumn{9}{c}{Percent of weights removed (\%)}\cr
    \cmidrule(lr){3-9}
    &&0	&68.37	&89.80 				&94.80	&98.66\cr
    \midrule
    Vanilla fine-tuning& w/o KD &94.17	&93.27			&91.67		&86.03	&54.39\cr
    Ours&w/o KD&94.17	&93.25			&91.83		&86.28	&54.92\cr  
    Ours &w/ KD&94.17	&\textbf{93.31}	&\textbf{92.07}		&\textbf{87.97}	&\textbf{56.88}\cr
    \bottomrule
    \end{tabular}
    \caption{Ablation study on  similarity-preserving knowledge distillation. }
    \label{tbl:ab_distall}
    }
\end{table}

\noindent \textbf{Expansion rate.} 
% Controlling the expansion weight $r \geq 1$ is the key to ensuring the existence of a matrix’s left and right inverse. Moreover, a large expansion rate can complicate the network and decrease the training speed. Therefore, 
We investigated the contribution of expanding the fully-connected layers in our experiments, and the results are presented in Tab.~\ref{ab_expansionrate}. A suitable expansion rate, such as 3, can enhance the overall performance of the network. Conversely, a large value, such as 5,  may overly complicate the network and hinder convergence.
\begin{table}[htbp]  
\centering
 \setlength{\tabcolsep}{1.5mm} \begin{tabular}{ccccc}
\toprule
Rate &Expand FC&FLOPs &Params &Acc(\%)\cr
\midrule
0&	w/o Expand FC &	43.298M &	215.642K &	88.91\cr
2&	w Expand FC &	313.736M &	1.675M &	89.43\cr
3&	w/o Expand FC &	591.660M&	3.127M&	\textbf{90.15}\cr
3&	w Expand FC &	591.665M&	3.132M&	90.12\cr
5&	w/o Expand FC &	1.392G&	7.282M&	90.07\cr
5&	w Expand FC	& 1.392G&	7.293M&	90.13\cr
\bottomrule
\end{tabular}
\caption{Expand a pruned MobileNetV2 applying ADMM strategy which has removed 92.18\% weights on CIFAR-10. Rate = 0 denotes the Vanilla fine-tuning without expansion.}
\label{ab_expansionrate}
% \vspace{-4mm}
\end{table}

\section{CONCLUSION}
\label{ssec:subhead}
% address的issue写成和前面强调的challenge一致
% In this paper, we introduce a linear over-parameterization method to address the issue of the difficulty of restoring the accuracy in structurally pruned neural networks.
% In this paper, we introduce a linear over-parameterization based pruning method to address the issue of the difficulty of restoring the accuracy in structurally pruned neural networks.
In this paper, we propose a linear over-parameterization approach to improve accuracy restoration in structurally pruned neural networks.
% Our approach expands convolutional/linear layers with consecutive layers via matrix decomposition without modifying the output feature maps to handle this issue . 
% In addition, we use similarity-preserving knowledgeto guide the fine-tuning of the over-parameterized network. 
It's achieved by fine-tuning the over-parameterized layers with increased parameters of the compact networks, during which similarity-preserving knowledge is exploited to maintain the feature extraction ability.
% The experimental results on CIFAR-10 and ImageNet datasets show that our fine-tuning method outperforms fine-tuning, especially for large pruning ratios. 
Comprehensive comparisons of CIFAR-10 and ImageNet datasets show that our approach outperforms fine-tuning, especially with large pruning ratios.
% Our method is focused on the fine-tuning stage of pruning and is complementary to other research areas such as pruning optimization, weight importance evaluation, and adaptive pruning ratios. Therefore, it can be applied simultaneously to these methods to improve the results.
Our method benefits the fine-tuning stage of pruning and synergizes with techniques in the areas of pruning optimization, weight importance evaluation, and adaptive pruning ratios. Consequently, it can be concurrently applied with these methods to enhance the overall effectiveness of pruning.
\bibliographystyle{IEEEbib}
\bibliography{strings,refs}

\end{document}